# DO PREDICTABILITY FACTORS TOWARDS SIGNING AVATARS HOLD ACROSS CULTURES?

*Abdelhadi Soudi*
*Sign Language Lab, Center for Languages and Communication,*
*ENSMR Rabat, Morocco*

*Manal El Hakkaoui*
*Sign Language Lab, Center for Languages and Communication, ENSMR Rabat, Morocco*

*Kristof Van Laerhoven*
*University of Siegen*
*Siegen, Germany*

## ABSTRACT

Avatar technology can offer accessibility possibilities and improve the Deaf and Hard-of-Hearing sign language users' access to communication, education and services, such as the healthcare system. However, sign language users' acceptance of signing avatars as well as their attitudes towards them vary and depend on many factors. Furthermore, research on avatar technology is mostly done by researchers who are not Deaf. The study examines the extent to which intrinsic or extrinsic factors contribute to predict the attitude towards avatars across cultures. Intrinsic factors include the characteristics of the avatar, such as appearance, movements and facial expressions. Extrinsic factors include users' technology experience, their hearing status, age and their sign language fluency. This work attempts to answer questions such as, if lower attitude ratings are related to poor technology experience with ASL users, for example, is that also true for Moroccan Sign Language (MSL) users? For the purposes of the study, we designed a questionnaire to understand MSL users' attitude towards avatars. Three groups of participants were surveyed: Deaf (N=57), Hearing (N=20) and Hard-of-Hearing (N=3). The results of our study were then compared with those reported in other relevant studies.

## 1. INTRODUCTION

Sign languages are visuospatial. The Deaf must rely on vision to develop language. Visual principles affect how that language is organized and expressed. Expression tends to be primarily manual and facial, and sign languages incorporate many techniques that visually and kinesthetically convey life's experiences. Unfortunately, unlike spoken languages, there are no conventionally used written coding systems to describe the elements of sign languages. As a result, to express a sign language in an ICT system, it must be depicted through graphics, animation, or video. There have been a few attempts to develop a written form to describe sign language (e.g., SignWriting (Sutton [17]) and Hamburg Notation System or HamNoSys (Prillwitz et al. [14]), but these are still hardly used by Deaf people or their service providers.

The use of signing avatars can facilitate content accessibility for the Deaf and Hard-of-Hearing. Unlike videos of human signers, animated signs represent signs anonymously. In addition, they can be modified at any time, and require far less bandwidth than videos. However, the development of virtual signers poses several challenges with regard to their quality and acceptability by SL users. The approval of signing avatars by SL users depend on several factors and has been discussed in several studies (Quandt et al. [15]; Kacorri et al. [9]; Kipp et al. [10]; Adamo-Villani [1]; Lu and Huenerfauth [13], [12]. Generally speaking, the results of these studies show that intrinsic factors (appearance and non-manual markers, such as facial expression, eye gaze and movement) are very important for SL users. Extrinsic factors, such as age of SL acquisition, technology experience and hearing status can also predict the attitude towards signing avatars.

For the purposes of this work, we designed a questionnaire to understand MSL users' general attitude towards avatars as well as towards the avatars used for the study. In order for us to compare the results of our study across different studies, we used as much as possible similar demographic and technology experience data reported about the participants of the relevant studies. More specifically, our study draws from, inter alia, the stimuli and evaluation questions released by Huenerfauth and Kacorri [6]. For the sake of an objective study of the attitude towards signing avatars, our questionnaire distinguishes between the attitude towards "an ideal signing avatar" and the attitude towards the avatar used in the study.

## 2. SURVEY AND RESPONDENTS

The goal of our study is to determine whether demographic factors and technology experience factors can affect the attitude of SL users towards avatars, and whether these factors have an impact on the participants' subjective judgements as well as their comprehension of the signing avatar used for this study. For this purpose, we designed a questionnaire that includes:

(i). Demographic Data: age, gender, hearing status, age of MSL acquisition
(ii). Technology experience: respondents' perception of the complexity of computers and their experience with smart phones and internet
(iii). Attitude towards "an ideal avatar": the participants are asked about their attitude towards the use of "ideal" avatars (that sign exactly as humans do) in real-world contexts, such as news and weather broadcast and in situations where privacy can be of concern
(iv). Comprehension rating of the avatar used in the study
(v). Naturalness
(vi). Impressions: the attitude towards the avatar's movement, facial expressions, handshape and location
Three groups of participants were surveyed: Deaf (N=57), Hearing (N=20) and Hard-of-Hearing (N=3). Compared to similar in-person studies (Kacorri et al. [9];[8]), our group of respondents is quite large.

## 3. METHOD, TASK AND STIMULI

Our study was conducted in-person. In an attempt to make sure that our research method is Deaf-friendly and adapts to the sociocultural experience of the respondents, two Deaf assistants (a female and a male) were recruited to explain the questions and the rating procedure on a paper questionnaire to each participant. Because of the cultural sensitivity of certain questions, the female Deaf assistant conducted the survey with the females and the male with the male participants. The importance of involving and how to involve Deaf assistants in the process of data collection and experimental studies has been documented in the literature (Huenerfauth et al. ([7]; Harris et al. [4]; Ladd [11]; Singleton et al. [16]). These authors stress the importance of, inter alia, making sure that the informed consent is translated in the native language of the respondents. This recommendation is particularly necessary in Morocco where the Deaf community has a very limited proficiency in spoken language.
After providing demographic information, their general attitude towards avatars, and technology experience, the participants rated their comprehension and impression of the signing avatars used in the study. For this purpose, seven individual MSL signs and one MSL sentence were used as the stimuli for the study: SICK, DOCTOR, MOTHER, FATHER, BABY, 1985, and THE MOTHER THE BABY FEEDS.
The avatars we created for the study are based on two different technologies. The first type of avatar we created is based on an enhanced version of HamNoSys notations.
The second type of avatar was created using the Blender environment.

| 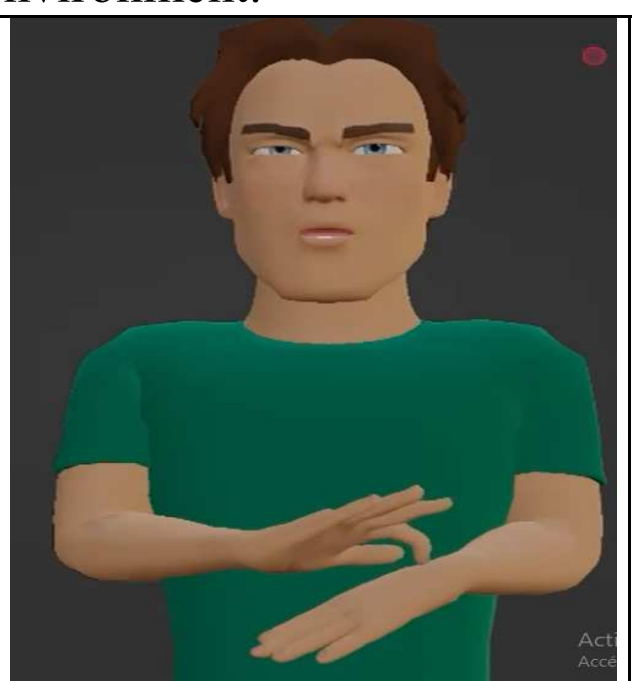 | 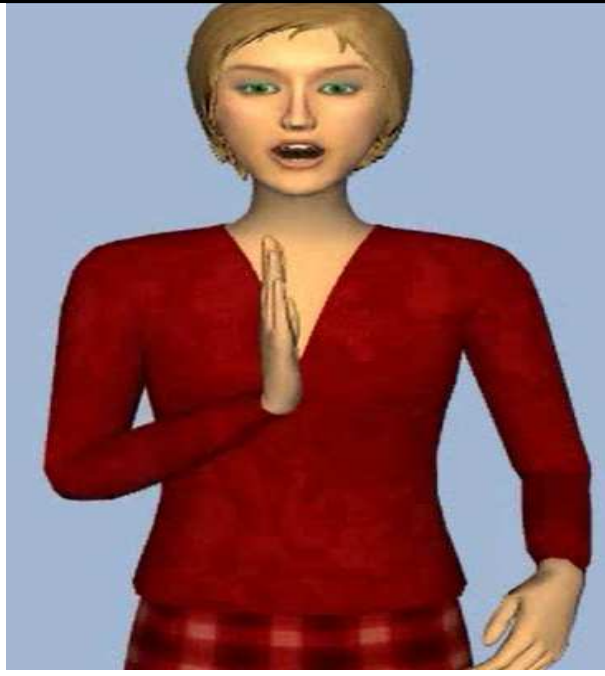 |
| --- | --- |
| The animation of the sign "SICK" in MSL using the Blender environment | The animation of the sign "FATHER" in MSL using an enhanced version of HamNoSys notations |

Using a 1-to-5 scalar response, the participants rated the comprehensibility of the two signing avatars by responding to the following question:
It was easy for me to understand the avatar's signing of the 7 words and sentence
كان الأفتار سهلاً بالنسبة لي لفهم كل من الكلمات السبع والجملة
We calculated the Comprehension variable by averaging the participants' answers.
The participants were then asked to answer questions measuring their subjective impression of the animation of the signs used in this study. They rated their responses to the questions below on a 5-point Likert scale:
1.The avatar's movements are natural حركات الأفتار مناسبة
2.The avatar's facial expressions are natural and correct تعابير وجه الأفتار مناسبة و صحيحة
3.The avatar uses natural and correct body language
يستخدم الأفتار لغة الجسد المناسبة
4.The avatar uses correct hand shapes
يستخدم الأفتار شكل اليد بطريقة صحيحة

## 4. ANALYSIS AND RESULTS

The purpose of this study is to examine whether demographical and experiential variables can predict the attitude of Sign language users towards avatars. For this purpose, we did multiple regression to analyze the data. Our results are compared with results from other relevant studies. First, we explored whether the age of acquisition of MSL can predict the attitude towards avatars, in general, and the participants' attitudes towards the avatar used in the study. We did a predictive model by using linear regression models based on the general attitude towards avatars and as independent variables the age of MSL acquisition and other variables relevant to our model (the hearing status, technology experience, age of participants, and gender). Our dependent variables are the general attitude towards avatars and the attitude towards the avatar used in the study with respect to comprehension, impression, and naturalness.

As can be seen in Table 1, the prediction of the general attitude of the Deaf is less positive than that of the hearing if we use the age of acquisition of MSL variable. However, with regard to the prediction of the attitudes towards the avatar used in the study, we notice that the Deaf are much more positive with the value $R^2$ (.064) lower for the hearing. The values of $R^2$ explains how much of the total variation in the dependent variable - General attitude or the attitude towards our avatar- can be explained by the independent variables (age of MSL acquisition and hearing status) as is illustrated in Table 2.

| ***Hearing Status*** | ***R*** | ***R Square*** | ***Adjusted R Square*** | ***Std. Error of estimation*** |
|---|---|---|---|---|
| Deaf | ,026[a] | ,001 | -,017 | 1,00885 |
| Hard of Hearing | ,336[a] | ,113 | ,004 | 1,23934 |
| Hearing | ,994[a] | ,988 | ,064 | ,30861 |

a. Predictors: (Constant), Age of Acquisition of MSL

**Table 1: General attitude towards avatars**

| ***Hearing Status*** | ***R*** | ***R Square*** | ***Adjusted R Square*** | ***Std. Error of estimation*** |
|---|---|---|---|---|
| Deaf | ,194[a] | ,038 | ,020 | 2,148 |
| Hard of Hearing | ,240[a] | ,113 | ,004 | 1,23934 |
| Hearing | ,081[a] | ,007 | ,049 | ,119 |

a. Predictors: (Constant), Age of Acquisition of MSL

**Table 2: The participants' attitudes towards the avatar used in the study**

We also examined the correlation between the participants' impression towards avatars and age of acquisition by running the multiple regression on this model. We noticed that the age of acquisition can play a significant role in the impression of the participants' attitudes towards avatars, especially for the Deaf participants (p value = ,848) as is shown in the ANOVA Table 3:

| Hearing status | Model | Sum of squares | Ddl | Mean square | F | Sig |
|---|---|---|---|---|---|---|
| Deaf | Regression | ,038 | 1 | ,038 | ,037 | ,848[b] |
| Hard of hearing | Regression | 3,525 | 1 | 3,525 | 2,295 | ,147[b] |
| Hearing | Regression | 8,127 | 1 | 8,127 | 85,333 | ,069[b] |

**Table 3: ANOVA table of the impression and age of acquisition variables**

To determine if significant correlations exist between the age of MSL acquisition and comprehension, we calculated the correlation coefficient using the Pearson method. As is shown in Table 4, the Age of Acquisition of MSL was found to be negatively correlated with comprehension with a value of (-.158) for the Deaf participants, (-.327) for the Hard-of-Hearing, and -,202 for the hearing, which implies that as the age of acquisition increases, comprehension decreases. Opposite results were reported by Quandt et al. [15].

| | Hearing status | | Age of acquisition | It was easy for me to understand the avatar |
|---|---|---|---|---|
| Deaf | Age of acquisition of MSL | Pearson correlation | 1 | -,158 |
| | | Sig. (bilateral) | | ,239 |
| | | N | 57 | 57 |
| | It was easy for me to understand the avatar's signing | Pearson correlation | -,158 | 1 |
| | | Sig. (bilateral) | ,239 | |
| Hard of Hearing | Age of acquisition of MSL | Pearson correlation | 1 | -,327 |
| | | Sig. (bilateral) | | ,788 |
| | | N | 3 | 3 |
| | It was easy for me to understand the avatar's signing | Pearson correlation | -,327 | 1 |
| | | Sig. (bilateral) | ,788 | |
| Hearing | Age of acquisition of MSL | Pearson correlation | 1 | -,202 |
| | | Sig. (bilateral) | | ,393 |
| | | N | 20 | 20 |
| | It was easy for me to understand the avatar's signing | Pearson correlation | -,202 | 1 |
| | | Sig. (bilateral) | ,393 | |

**Table 4: Correlations between comprehension and the age of acquisition**

With respect to the correlation between the general attitude towards avatars and Technology experience, the results show that there is a positive correlation between the technology experience and the general attitude with a value of (,263) for the Deaf as is shown in table 5. This means that Deaf participants who have less technology experience are likely to have negative attitude towards avatars. Similar results were reported by Kacorri et al. [8].

| | Hearing status | | General attitude | Tech experience |
|---|---|---|---|---|
| Deaf | General attitude | Pearson correlation | 1 | ,263* |
| | | Sig. (bilateral) | | ,048 |
| | | N | 57 | 57 |
| | Tech experience | Pearson correlation | -,263* | 1 |
| | | Sig. (bilateral) | 048 | |
| | | N | 57 | 57 |
| Hearing | General attitude | Pearson correlation | 1 | ,068 |
| | | Sig. (bilateral) | | ,776 |
| | | N | 20 | 20 |
| | Technology experience | Pearson correlation | ,068 | 1 |
| | | Sig. (bilateral) | ,776 | |
| | | N | 20 | 20 |

**Table 5: correlations between the general attitude and the technology experience**

As is shown in Table 6, the attitude of the Deaf is less positive than that of the hearing with respect to the technology experience variable. This can be explained by the fact that the technology experience of the hearing participants is higher (.986) than that of the Deaf participants (.052). Access to technology by the Deaf can be explained by the high rate of illiteracy of the Deaf community in Morocco where more than 85% of Deaf children do not have access to education.

| Hearing status | Model | R | R Square | Adjusted R square | Std. Error of estimation |
|---|---|---|---|---|---|
| Deaf | 1 | ,263[a] | ,069 | ,052 | ,97361 |
| Hard of hearing | 1 | ,068[a] | ,005 | -,051 | 1,31292 |
| Hearing | 1 | ,997[a] | ,993 | ,986 | ,23570 |

a. Predictors: (Constant), Technology Experience

**Table 6: Multiple regression of the general attitude towards the avatar based on technology experience**

It is worth noting that when we calculate the prediction of the general attitude by using just the demographic variables (age of participants, age of acquisition, and gender) the value of $R^2$ is ,306, which is lower compared to when we also include the technological variable ( $R^2$ is ,315) as is shown in Tables 7 and 8. Similar results were reported by Kacorri et al. [8] through a survey of 62 participants.

| Model | R | R sq. | Adjusted R sq. | Std. Error of estimation |
|---|---|---|---|---|
| 1 | ,591[a] | ,350 | ,306 | 1,10817 |

a. Predictors: (Constant), Hearing Status, Technology Experience, Gender, Age of Acquisition of MSL, age of the participants

**Table 7: Multiple regression model of general attitude with only demographical variables**

| Model | R | R sq. | Adjusted R sq. | Std. Error of estimat. |
|---|---|---|---|---|
| 1 | ,591[a] | ,350 | ,315 | 1,10076 |

a. Predictors: (Constant), Hearing Status, Technology Experience, Gender, Age of Acquisition of MSL.

**Table 8: Multiple regression model of general attitude with demographical and technological variables**

That is, the more we add more significant values to the model, the higher the R squared value becomes, and as a result, the prediction is higher. It is also noticed that there is a correlation between Technology experience and the age of participants. Technology experience decreases (-.124) when the age of participants increases, which means that people who are younger are more positive and willing to use technology than older participants (cf. Table 9).

| | | Tech experience | Participants' age |
|---|---|---|---|
| Technology experience | Pearson correlation | 1 | -,124 |
| | Sig. (bilateral) | | ,274 |
| | N | 80 | 80 |
| Age of the participants | Pearson correlation | -,124 | 1 |
| | Sig. (bilateral) | ,274 | |
| | N | 80 | 80 |

**Table 9: correlations between the technology experience and the age of participants**

## 5. CONCLUSION AND RECOMMENDATIONS FOR FUTURE WORK

In this work, we presented the results of a study that examined the extent to which demographical and technological factors contribute to predict the attitude of MSL users towards virtual signers. It was shown that there is a positive correlation between technology experience and the general attitude towards avatars. That is, participants who have less technology experience are likely to have negative attitude towards avatars. That is, participants who have less technology experience are likely to have negative attitude towards avat.ars. The same result has been reported in prior work. Our study also supports prior work that has shown that the avatars that appeal most to SL users are those with better non-manual markers. However, it is worth noting that, contrary to what was reported in prior work, the Age of Acquisition of MSL was found to be negatively correlated with Comprehension, Naturalness and Impression. The results show that as the age of acquisition increases, Comprehension, Naturalness, and Impression ratings decreases.

In view of the high degree of regional variation in MSL and the different cultural backgrounds of sign language users, it is recommended that future work collect large datasets from users rating avatars. Interestingly, there are mainly two ethnolinguistic groups in Morocco: the Arabs and the Amazighs. We believe that understanding the sociolinguistic situation of MSL by investigating factors, such as the multilingual linguistic environment, gender, regional variation, family and education can add another dimension for research that examines the factors that can predict the attitude of sign language users towards avatars.

In order to compare results across different studies, we, following Kacorri et al. [8], strongly encourage researchers to use similar standard survey questions.